\pdfoutput=1

\documentclass[11pt]{article}

\usepackage[dvipsnames]{xcolor}

\usepackage{EMNLP2022}

\usepackage{times}
\usepackage{latexsym}

\usepackage[T1]{fontenc}
\usepackage[utf8]{inputenc}

\usepackage{microtype}

\usepackage{inconsolata}
\usepackage{times}
\usepackage{latexsym}
\usepackage[pdftex]{graphicx}
\usepackage{adjustbox}
\usepackage{xspace}
\usepackage{breqn}
\usepackage{amsmath}
\usepackage{enumitem}
\usepackage{longtable}
\usepackage{multirow}
\usepackage{array}
\usepackage{makecell}
\usepackage{tabularx}
\usepackage{booktabs}
\usepackage{amsthm}
\usepackage[english]{babel}
\usepackage{blindtext}
\usepackage{tablefootnote}

\newtheorem{lemma}{Lemma}

\newcommand{\colorann}[3]{\textcolor{#1}{${}^{#2}[$#3$]$}}

\newcommand{\jon}[1]{\colorann{red}{Jon}{#1}}
\newcommand{\justin}[1]{\colorann{blue}{Justin}{#1}}
\newcommand{\ab}[1]{\colorann{cyan}{Ahmad}{#1}}
\newcommand{\chris}[1]{\textcolor{red}{\textbf{*Chris*}: #1}}

\renewcommand{\jon}[1]{}
\renewcommand{\justin}[1]{}
\renewcommand{\ab}[1]{}
\renewcommand{\chris}[1]{}

\newcommand{\ours}{PrefineDST\xspace}
\newcommand{\ourmetric}{CheckDST\xspace}

\definecolor{skyblue}{RGB}{194,237,255}
\definecolor{orange}{RGB}{255,204,145}

\usepackage{microtype}
\usepackage{todonotes}

\title{Know Thy Strengths: \\ Comprehensive Dialogue State Tracking Diagnostics}

\author{
Hyundong Cho$^{1,2}$\thanks{\xspace~\xspace~The work of HC and AB was done at Meta AI when HC was an intern. This work was done prior to JM joining Amazon.} \, Chinnadhurai Sankar$^2$ \, Christopher Lin$^2$ \, Kaushik Ram Sadagopan$^2$\\
\textbf{Shahin Shayandeh$^2$ \, Asli Celikyilmaz$^2$ \, Jonathan May$^1$ \, Ahmad Beirami$^{3*}$} \\
$^1$University of Southern California, Information Sciences Institute \, $^2$Meta AI \, $^{3}$Google Research \\
\small{\texttt{hd.justincho@gmail.com}}
}

\begin{document}
\maketitle

\begin{abstract}

Recent works that revealed the vulnerability of dialogue state tracking (DST) models to distributional shifts have made holistic comparisons on robustness and qualitative analyses increasingly important for understanding their relative performance. 
We present our findings from standardized and comprehensive DST diagnoses, which have previously been sparse and uncoordinated, using our toolkit, \ourmetric, a collection of robustness tests and failure mode analytics. 
We discover that different classes of DST models have clear strengths and weaknesses, where generation models are more promising for handling language variety while span-based classification models are more robust to unseen entities.  
Prompted by this discovery, we also compare checkpoints from the same model and find that the standard practice of selecting checkpoints using validation loss/accuracy is prone to overfitting and each model class has distinct patterns of failure.   
Lastly, we demonstrate how our diagnoses motivate a pre-finetuning procedure with non-dialogue data that offers comprehensive improvements to generation models by alleviating the impact of distributional shifts through transfer learning.

\end{abstract}

\section{Introduction}

\begin{figure}[t]
    \centering
    \includegraphics[width=0.9\columnwidth]{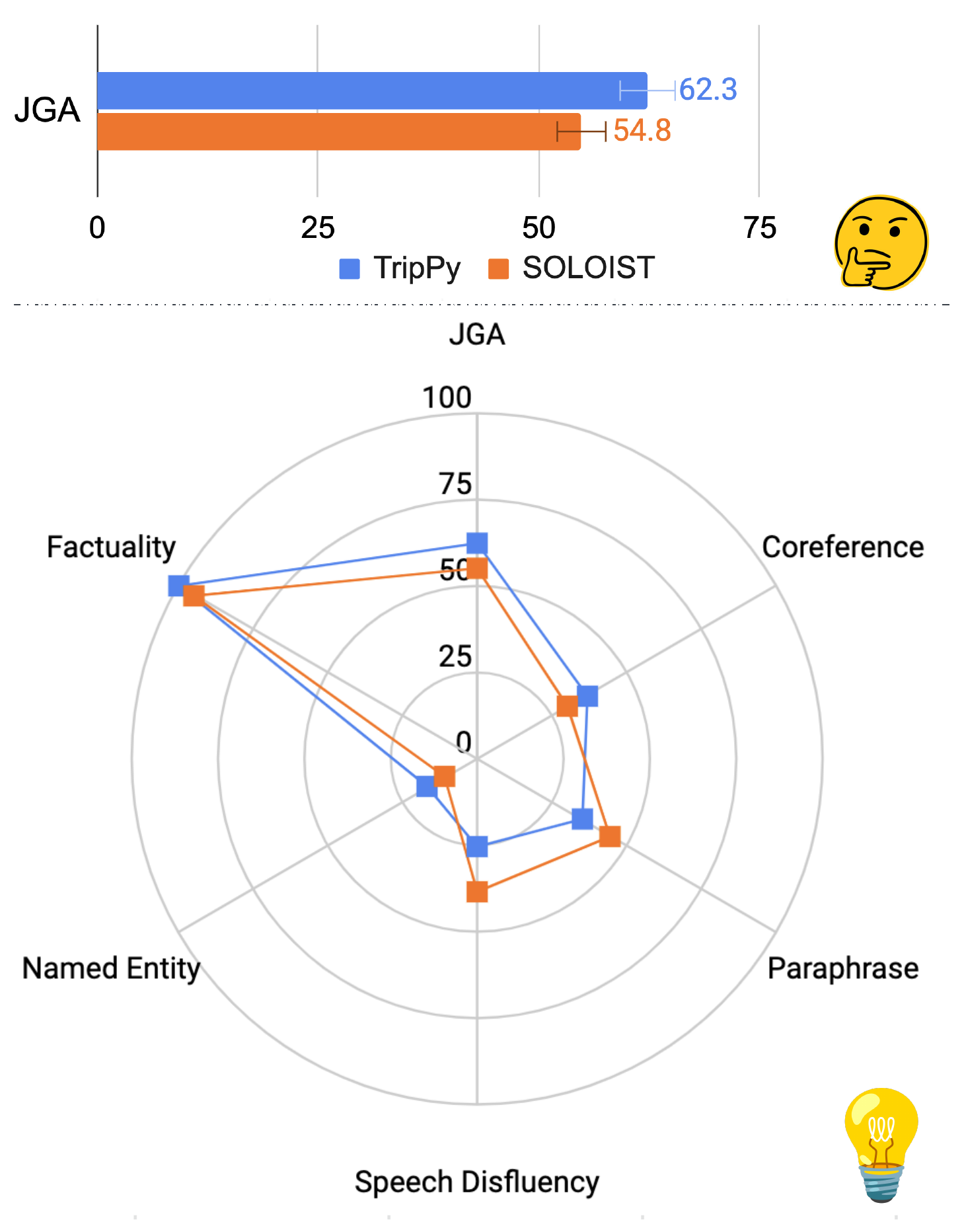}
    \caption{
            \textit{Top:} performance differences on current benchmarks provide limited information on how DST models compare in robustness and various modes of failure. \textit{Bottom:}  \ourmetric diagnoses paints a richer summary that highlights strengths and weaknesses through standardized and comprehensive comparisons of metrics that isolate various perturbations and failure modes. Higher is better for all metrics.  
        }
    \label{fig:intro_example}
    \vspace{-.15in}
\end{figure}

A crucial skill for task-oriented dialogue models, which serve as backbones to modern digital assistants, is slot filling, formally known as dialogue state tracking (DST).
It requires understanding the users' intents to populate slots in API queries that request information needed to fulfill their goals.
To encourage the development of DST models, various large-scale datasets that double as benchmarks have been developed, such as MultiWOZ \cite{budzianowski2018large}, Taskmaster \cite{ byrne-etal-2019-taskmaster}, and the Schema-guided Dialogue (SGD) dataset \cite{rastogi2020towardssgd}.

As shown in the bar chart on the top side of \autoref{fig:intro_example}, recent state-of-the-art approaches in DST leaderboards such as MultiWOZ \cite{budzianowski2018large} are compared to one another on a single metric that provides limited information and their modes of failure. 
Moreover, since these models demonstrate a sharp drop in joint goal accuracy (\texttt{JGA}), the average accuracy of correctly predicting all slots for a dialogue turn, when exposed with realistic distributional shifts \cite{li2020coco, qian-etal-2021-annotation, liu-etal-2021-robustness, peng-etal-2021-raddle}, it is increasingly important to shift the focus from answering the question \textit{``\textbf{Is} one DST model better than another?''} to \textit{``\textbf{How} does one DST model compare to another?''}
through robustness tests and qualitative analyses to understand relative performance. 
Yet these efforts have mostly been sparse and uncoordinated, precluding in-depth comparisons and opportunities to discover improvements that combine strengths from different models.

To address this gap, we first consolidate previous efforts in robustness testing and qualitative analyses to design our toolkit, \ourmetric\footnote{Our code and data are available at \url{https://github.com/wise-east/CheckDST}.}.
\ourmetric facilitates comprehensive diagnosis by quantifying robustness with augmented test sets that represent various perturbations in isolation and measuring the frequency of common failure modes. 
For measuring robustness, we introduce \textit{consistency} JGA (\texttt{cJGA}) to place more emphasis on prediction consistency and obviate the assumption that the perturbations should be more difficult than their corresponding original samples.

With a \ourmetric diagnosis, we can quickly examine a rich comparison summary that highlights relative performance in multiple dimensions, as illustrated in the bottom of \autoref{fig:intro_example}. 
Evaluating a subset of models from two major classes of state-of-the-art models, span-based classification models and SimpleTOD-style generation models (henceforth denoted as classification models and generation models, respectively), on the MultiWOZ \cite{budzianowski2018large} leaderboard, we demonstrate the value of our diagnosis through the revelation that robustness is not proportionate to higher \texttt{JGA} and that the model classes have clearly different strengths and weaknesses. 
In particular, while classification models attain higher \texttt{JGA}, generation models are significantly more robust to various perturbations except to swapped named entities.
This disconnect between JGA and robustness found in inter-model comparisons naturally prompts comparisons between different checkpoints from the same model. 
By diagnosing multiple intermediate checkpoints with \ourmetric, we show that the current standard practice of selecting a checkpoint according to performance on the validation set is prone to overfitting, regardless of model class. 
\justin{see if there is a way to suggest an alternative to the standard practice that works as a good rule of thumb}
Finally, we demonstrate how the findings from our diagnoses with \ourmetric can help develop more robust DST models with strategies that alleviate the trade-off between \texttt{JGA} and robustness and mitigate the failure modes. 
We share our experiments with \ours, building on the robustness of generation models via a pre-finetuning procedure with instruction prompts, similarly to the T0 model \cite{sanh2021multitask}, with non-dialogue tasks to acquire skills that should intuitively boost robustness. 
We show that comprehensive improvements that reduce the trade-off between \texttt{JGA} and robustness can be achieved without directly exposing the model to similar perturbations used in the test sets during training. 

In summary, our contributions include: 
\vspace{-.05in}
\begin{itemize}
    \item \ourmetric, a toolkit for facilitating a standardized and comprehensive diagnosis of DST models; \vspace{-.1in}
    \item a rich empirical study enabled by \ourmetric that discovers the disconnect between \texttt{JGA} and robustness, that major classes of DST models have clear strengths and weaknesses that is not portrayed in the performance difference on the original test set, and that current checkpoint selection criteria are prone to overfitting; and 
    \vspace{-.1in}
    \item \ours, a simple model motivated by \ourmetric diagnosis that achieves a comprehensive improvement in robustness and \texttt{JGA} through a multitasking pre-finetuning step with instruction-prompts. 
\end{itemize}

\section{DST Diagnostics with \ourmetric}
\label{checkdst}

\begin{table*}[t]
\small
    \begin{adjustbox}{width=\textwidth,center}

\begin{tabular}{@{}llll@{}}
\toprule
    \multicolumn{1}{l}{} & \multicolumn{2}{c}{Input Example}  & \multicolumn{1}{c}{Example DST Prediction} \\ \midrule

    \multicolumn{1}{l}{\texttt{Original} } & \multicolumn{2}{l}{\textit{I would like to leave from Cambridge.}}  & \multicolumn{1}{c}{\texttt{train departure \textcolor{Green}{cambridge}}} \\ 
    
    \multicolumn{1}{l}{\texttt{Paraphrase} } & \multicolumn{2}{l}{\textit{\textcolor{blue}{Please book me one ticket departing} from Cambridge. }}  & \multicolumn{1}{c}{\texttt{train departure \textcolor{Green}{cambridge}}} \\ 
    
    \multicolumn{1}{l}{\texttt{Speech Disfluency} } & \multicolumn{2}{l}{\textit{I would like to \textcolor{blue}{uh} leave from \textcolor{blue}{London no I meant} Cambridge.}}  & \multicolumn{1}{c}{\texttt{train departure \textcolor{Green}{cambridge}}} \\

    \multicolumn{1}{l}{\texttt{Unseen Entity} } & \multicolumn{2}{l}{\textit{I would like to leave from \textcolor{blue}{mbadgceir}.}}  & \multicolumn{1}{c}{\texttt{train departure \textcolor{Green}{mbadgceir}}} \\    
    
    \midrule
    
    \multicolumn{1}{l}{\multirow{1}{*}{\texttt{Coreference}}}  & 
    
    \multicolumn{2}{l}{\begin{tabular}[c]{@{}l@{}}
\textit{\makecell[l]{I need you to \underline{book the restaurant} for \underline{Tuesday.} \\ \texttt{...}\\  %
I'm also looking for a train to London Kings Cross \\~on the \underline{same day as the restaurant booking.}}}
\end{tabular}} 
    
    & \texttt{\makecell[l]{...\\restaurant day tuesday\\train day \textcolor{Green}{tuesday}\\...}} \\ 
    
    \midrule

    \multicolumn{1}{l}{\texttt{Hallucination} } & \multicolumn{2}{l}{\textit{I would like to leave from \textcolor{blue}{London}.}}  & 
    
    \multicolumn{1}{l}{\begin{tabular}[c]{@{}l@{}}
\textit{\makecell[l]{
\texttt{train departure \textcolor{red}{cambridge}} \\
\texttt{train departure \textcolor{Green}{london}}
}}
\end{tabular}}

    \\

\bottomrule
\end{tabular}
    \end{adjustbox}
\caption{An overview of perturbations and failure modes captured in \ourmetric illustrated with simple examples. \textcolor{blue}{Blue} segments indicate changes from the original input. Dialogue states in \textcolor{Green}{green} are correct predictions, while those in \textcolor{red}{red} are incorrect. 
\justin{reorganize and add what is meant by hallucination}}
    \vspace{-.1in}
\label{tab:ourmetric_overview}
\end{table*}

To conduct standardized robustness tests and qualitative analyses, we first consolidate previous efforts to design 
\textit{Checklist for Dialogue State Tracking} (\ourmetric), a toolkit for quantifying robustness with a collection of augmented test sets and diagnostics that help identify commonly known failure modes. 
In this section, we introduce \textit{consistent} \texttt{JGA} (\texttt{cJGA}) and summarize the perturbations and failure modes of interest that are shown in \autoref{tab:ourmetric_overview}, as well as their corresponding metrics.

\subsection{Measuring counterfactual robustness with Consistent \texttt{JGA} (\texttt{cJGA})}
\label{sec:cJGA}

By exposing DST models to perturbations, we want to quantify DST models' responses to valid perturbations that may be encountered at deployment in order to answer the question \textit{``How do their robustness compare to other models?''}
In this work, we define \textit{robustness} as the ``degree of performance consistency to realistic distribution shifts.''

Based on this definition, robustness can be measured by comparing \texttt{JGA} to \texttt{JGA} on a perturbed test set ($\widetilde{\texttt{JGA}}$), but this requires the perturbed test set to strictly contain corresponding samples that are more difficult for the model such that the performance drop represents a lack of robustness.
There may be cases where certain perturbed samples are easier than the original, leading a model to achieve $\widetilde{\texttt{JGA}}$ similar to \texttt{JGA}, even though it makes many {\em inconsistent} predictions between the original and perturbed pairs. 
Therefore, we measure robustness through \textit{consistent} \texttt{JGA} (\texttt{cJGA}) to obviate the difficulty requirement of the perturbations.  

\texttt{cJGA} simply measures the frequency of the cases where the prediction is correct on both the original and perturbed samples.
Given a DST model (with parameters $\theta$), let function $f(z; \theta) \to \{0, 1\}$ indicate whether the joint goal is satisfied on sample $z = (x, y)$, where $x$ is the dialogue history and $y$ is the reference belief state. Further, let $\widetilde{z} = (\widetilde{x}, \widetilde{y})$ denote a perturbed sample (e.g., with paraphrased dialog history). Then, we define \texttt{cJGA} for a sample set $[n]:= \{1, \ldots, n\}$ as:
\begin{equation}
    \label{conditional_jga}
    \texttt{cJGA} := \frac{1}{n} \sum_{i \in [n]} \mathbf{1}(f(z_i; \theta) = f(\widetilde{z_i}; \theta) = 1),
\end{equation} 
where $\mathbf{1}(\cdot)$ denotes the indicator function.

\justin{add qualitative example to make the case for cJGA, + diagram/table how it is measured}

When labels are preserved, i.e. $y$ and $\widetilde{y}$ are identical, \texttt{cJGA} is an adaptation of the CheckList \textit{invariance} test, 
and if changes from $y$ to $\widetilde{y}$ are mirrored in changes from $x$ to $\widetilde{x}$, it is an adaptation of the Checklist \textit{directional} test \cite{ribeiro-etal-2020-beyond}.
We also make the mathematical case for the usefulness of \texttt{cJGA} by proving that $ \texttt{cJGA}  \leq \min\{ \texttt{JGA} , \widetilde{\texttt{JGA}} , 1 - |\texttt{JGA} - \widetilde{\texttt{JGA}}|\}$ in Lemma~\ref{lem:cJGA} (Appendix~\ref{app:cJGA}), with equality only if the model performance is \textit{consistent} on perturbed samples and original ones. This establishes that \texttt{cJGA} captures robustness beyond the JGA drop as it additionally captures the consistency of performance across the original and perturbed test set.

\subsection{Perturbations}

We select perturbations from those
suggested by \newcite{liu-etal-2021-robustness, peng-etal-2021-raddle, qian-etal-2021-annotation} and we elaborate them briefly here.\footnote{Note that the perturbations we select are relatively simple ones applied at the utterance level. 
There are previous works that create augmented test sets with entire new dialogues generated with counterfactual goals~\cite{li2020coco} or insert additional dialogue turns~\cite{peng-etal-2021-raddle}. 
While dialogue-level perturbation is interesting, its difficulty to decouple from other lower-level perturbations such as paraphrases and unseen named entities make it inadequate for isolated analysis and therefore we do not include it in \ourmetric.} 

\paragraph{Paraphrase.}

A robust DST model should make consistent predictions for utterances that have the same semantics, regardless of language variety. 
There is a wide spectrum for what is considered a paraphrase, including  single word replacements with synonyms.\footnote{According to \newcite{li2020coco}, DST models only drop 2\% in JGA for these kinds of simple paraphrases. However, when the paraphrases share only a few words with the original, the models demonstrate significant drops in JGA  \cite{peng-etal-2021-raddle, liu-etal-2021-robustness}, indicating that understanding paraphrases is still a challenge.} 
For \ourmetric, we focus on more complex paraphrases with minimal word overlap.  

In the context of DST, paraphrasing is defined as any change to the wording of utterances that preserves the dialogue belief states. 
Thus, Paraphrase Invariant \texttt{cJGA} (\texttt{PI cJGA}) measures whether a model can make correct slot predictions consistently for two semantically equivalent utterances. 

\paragraph{Speech Disfluency.}

Many task-oriented dialogue applications are built around voice-based digital assistants. 
Therefore, a DST model's resilience to speech artifacts is a crucial criterion of a TOD model's success. 
Speech disfluencies are common speech artifacts that include the restart of requests mid-sentence, use of non-lexical vocables or filler words, and stammering and repetition that occur within the flow of otherwise fluent speech \cite{wang2020multi}. 
As with \texttt{PI cJGA}, Speech Disfluency Invariant \texttt{cJGA} (\texttt{SDI cJGA})  measures how often a model maintains correct predictions with the presence of speech disfluencies.

\paragraph{Unseen Entities.}
DST models should be able to generalize performance to unseen entities, but it is a known problem that they can overfit to entities that appear frequently in the training set \cite{qian-etal-2021-annotation}. 
To measure their robustness to entities not seen during training, we replace named entities in the dialogue belief states and conversations with scrambled entities. 
Named Entity Directional \texttt{cJGA} (\texttt{NED cJGA}) tracks how frequently a model correctly mirrors a change in the conversation to its prediction to obtain the right slot values.

\justin{add dialogue perturbations (additional utterances that don't change DST, some examples in RADDLE already explored, but they are very unnatural insertions)}

\subsection{Failure modes}

\paragraph{Hallucination.}

Generation models, models that autoregressively generate a sequence of text from an open vocabulary, have become popular for task-oriented dialogue \cite{su2021pptod, peng2020soloist, hosseini2020simple} following their success with various NLP tasks, but they are known to suffer from content \textit{hallucination}, providing irrelevant entities memorized from training~\cite{massarelli-etal-2020-decoding, maynez-etal-2020-faithfulness}. 
Therefore, we measure hallucination frequency as well in \ourmetric with Factuality ($F$). $F$ is equal to $1$ if the predicted named entity is in the dialogue  history and $0$ otherwise.
Since hallucination occurs even without any perturbations, 
\ourmetric reports $F$  on both the original test set and one used for \texttt{NED cJGA} ({$F_{\texttt{orig}}$} and \texttt{$F_{\texttt{swap}}$}, respectively). 

\paragraph{Coreference resolution.}
Long conversations with coreferences that span multiple turns are especially challenging, as shown by the performance improvement when coreference annotations are present \cite{quan-etal-2019-gecor, han2020multiwoz23}. 
As a proxy for measuring a model's ability to understand longer conversations and resolve coreferences to make correct predictions, we simply calculate the \texttt{JGA} for samples in the original test set that require coreference resolution and denote it as \texttt{CorefJGA}.

\section{Experimental Setup}
\label{experiments}

\begin{table*}[ht!]
\begin{adjustbox}{width=\textwidth,center}
    \centering
    \begingroup
    \small
    \setlength{\tabcolsep}{6pt} %

    \begin{tabular}{clcccccccccccc} 
            
        \toprule
        & Model & \texttt{JGA} &  \texttt{CorefJGA}& \texttt{PI cJGA} & \texttt{SDI cJGA} & \texttt{NED cJGA} & $F_{\texttt{orig}}$ & $F_{\texttt{swap}}$ \\ \hline
        
        \multirow{2}{*}{SCLS} & TripPy \citeyearpar{heck-etal-2020-trippy} & \colorbox{Green}{$62.3_{0.2}$} & \colorbox{Green}{$37.0_{0.7}$} & \colorbox{pink}{$36.2_{0.1}$} & \colorbox{pink}{$27.7_{0.5}$} & \colorbox{Green}{$16.5_{0.4}$} & \colorbox{Green}{$100_0$} &  \colorbox{Green}{$100_0$} \\  [-0.5em]
        & ConvBERT-DG \citeyearpar{MehriDialoGLUE2020} & \colorbox{Green}{$62.0_{0.1}$} & \colorbox{Green}{$36.6_{0.7}$} & \colorbox{pink}{$35.9_{0.2}$} & $29.5_{0.4}$ & \colorbox{Green}{$16.2_{0.1}$} & \colorbox{Green}{$100_0$} &  \colorbox{Green}{$100_0$} \\ \hline

        \multirow{3}{*}{GEN} & BART-DST 
        \citeyearpar{lewis-etal-2020-bart}
        & \colorbox{pink}{$52.5_{0.2}$} & \colorbox{pink}{$25.5_{1.2}$} & $43.0_{0.8}$ & $36.5_{1.3}$ & $9.8_{0.7}$ & \colorbox{pink}{$94.8_{0.2}$} & $75.4_{1.8}$ \\ [-0.5em]

        & SOLOIST \citeyearpar{peng2020soloist} & $54.8_{0.4}$ & $30.4_{1.1}$ & \colorbox{Green}{$44.5_{0.6}$} & \colorbox{Green}{$38.5_{0.5}$} & $10.7_{0.4}$ & \colorbox{pink}{$94.9_{0.1}$} & $81.1_{1.6}$ \\ [-0.5em]
        & MUPPET-DST \citeyearpar{aghajanyan2021muppet} & $54.9_{0.4}$ & $29.9_{1.9}$ & \colorbox{Green}{$45.4_{0.4}$} & \colorbox{Green}{$39.1_{0.7}$} & \colorbox{pink}{$7.8_{1.4}$} & \colorbox{pink}{$94.6_{0.1}$} & \colorbox{pink}{$68.4_{3.8}$} \\

        \bottomrule

    \end{tabular}
    \endgroup 
\end{adjustbox}
    \caption{\ourmetric diagnosis overview on the MultiWOZ~2.3 dataset clearly shows a significant divergence of robustness properties between the two model classes. SCLS is short for span-based classification models while GEN is short for generation models. 
    All results are percentages aggregated over five runs with different seed values, presented as $\mathtt{median_{se}}$, where $\mathtt{se}$ is short for standard error. 
    \colorbox{Green}{ $\mathtt{x}$ } marks the best score for the column while \colorbox{pink}{ $\mathtt{x}$ } marks the worst. If there is an overlap between $\mathtt{median - se}$ and $\mathtt{median + se}$ with the best/worst score, the difference is considered statistically insignificant and all overlapping scores are highlighted.
    }
    \vspace{-.1in}
    \label{tab:full_shot_results}
\end{table*}

\subsection{DST Benchmark}
\label{sec:dataset}

We use MultiWOZ \cite{budzianowski2018large} as an example TOD dataset that we apply \ourmetric to. 
It is an open-source dataset released with the Apache 2.0 license and we use it for research purposes only.
We specifically use MultiWOZ~2.3 \cite{han2020multiwoz23}, which includes corrections from MultiWOZ~2.1 \cite{eric2020multiwoz21} and coreference annotations, and use its original train/dev/test splits.

\paragraph{Perturbation Tools.}
For augmented test sets with paraphrases and speech disfluencies, we use the augmented test sets from 
LAUG~\cite{liu-etal-2021-robustness}\footnote{Paraphrases are collected with a SC-GPT model \cite{peng-etal-2020-shot} trained to generate the user response given its corresponding dialogue history and the dialogue act. The resulting paraphrases are heuristically filtered to ensure that slot values in the belief states are preserved.
The degree of paraphrasing with LAUG is significant, replacing 74\% of all words. 
Speech disfluencies are inserted according to their occurrence frequency in the Switchboard corpus \cite{godfrey1992switchboard}. 
For both perturbations, more than 97\% were considered appropriate by human evaluators. 
Refer to \newcite{liu-etal-2021-robustness} for further details.} 
and update the dialogue states with those from the official MultiWOZ~2.3 dataset to resolve inconsistencies we found.
LAUG is an open-source augmentation toolkit that can be used for any task-oriented dialogue dataset that has dialogue acts and belief state annotations. 
To create a test set with unseen entities, we scramble the character order of named entity slot values~\cite{huang2021dair} to create unseen entities, as shown in \autoref{tab:ourmetric_overview}.\footnote{Instead, we can swap with more realistic entities not seen during training, such as those from Schema Guided Dialogue (SGD) \cite{rastogi2020towardssgd, qian-etal-2021-annotation}. 
However, since some baseline models are pre-trained with SGD, we choose scrambled entities as the default for a fair comparison. }

\justin{replace with realistic entities not seen in either MultiWOZ or SGD}

\subsection{Models}
\label{sec:baselines}

From the top-performing models reported on the MultiWOZ~2.0 repository
and the MultiWOZ~2.3 repository, we implement a subset that has replicable code.
We train all models for 10 epochs with the default hyperparameters reported in their corresponding reports, if applicable. 
We provide all other training details in Appendix \ref{app:baseline_training}.
For inter-model comparisons, we conduct \ourmetric diagnosis on checkpoints with the best validation \texttt{JGA}.

Recent DST models that attain competitive results can largely be divided into two classes: classification models and generation models. 

\paragraph{Classification models.} 
These models predict the required dialogue state operation and then specify the starting and ending index of slot values in the context to copy from or choose labels from a predefined ontology for those that are not directly in the context. 
The domains and their slot types are fixed, and predictions are made for every possible (\texttt{domain}, \texttt{slot-type}) pair using a classification layer.

\noindent\textbf{(\textit{i}) TripPy} \cite{heck-etal-2020-trippy} is a model based on BERT \cite{devlin-etal-2019-bert} that determines whether slot values can be copied from the current utterance, the previous system utterance, or the previous turns dialogue belief state.

\noindent\textbf{(\textit{ii}) ConvBERT-DG} \cite{MehriDialoGLUE2020} is TripPy with BERT replaced with ConvBERT-DG, a BERT model that is further pretrained on more than 70 million conversations of open-domain dialogue and then finetuned on the DialoGLUE benchmark.

\paragraph{Generation models.}
Generation models for DST predict belief states in the same way the underlying model generates text. 
They sequentially generate the \texttt{domain}, \texttt{slot-type}, and \texttt{slot-value}. 
Belief states are usually generated usually through greedy sampling on $P(x_t| x_{1:t-1}, C; \theta))$, where $X=\{x_1, x_2, ... x_t\}$ is the flattened text format of the belief state, e.g. \texttt{domain slot-type slot-value}, $C$ is the dialogue context, and $\theta$ is the set of model parameters.  
Generation models are becoming increasingly popular as they can easily be expanded to perform end-to-end task-oriented dialogue by also generating the dialogue policy and responses after the belief states.

\noindent\textbf{(\textit{i}) BART-DST}~\cite{lewis-etal-2020-bart} is a model that is trained in the SimpleTOD approach~\cite{hosseini2020simple} to generate the dialogue belief states in \texttt{domain} \texttt{slot-type} \texttt{slot-value} format given a conversation.

\noindent\textbf{(\textit{iii}) SOLOIST} \cite{peng2020soloist} is also similar to BART-DST, but it excludes dialogue act prediction during end-to-end training and adds a pre-training step with the SGD dataset. We use BART instead of GPT-2 in our experiments. 

\noindent\textbf{(\textit{iv}) MUPPET} \cite{aghajanyan2021muppet} is a BART-DST model that is pre-finetuned on more than 50 natural language tasks.
MUPPET adds auxiliary layers that take the representation of the final token in BART to perform classification tasks and does standard autoregressive language modeling for generation tasks. 

\section{\ourmetric Diagnosis Results}
\label{analysis}

\subsection{Inter-model comparison}     
\vspace{-.05in}

\paragraph{Each model class has distinct strengths and weaknesses.}
With our results in  \autoref{tab:full_shot_results}
we can immediately see a dramatic divergence of robustness properties between the classification and generation models.\footnote{We assess the generation models the same way as we would for classification models by including ``dontcare'' slots. This makes an enormous difference. Otherwise, \texttt{JGA} and \ourmetric metrics are boosted by more than 5\%.}
Overall, generation models are much more robust to language variety, as captured by paraphrases and speech disfluencies, while classification models are significantly more robust to unseen entities. 
That the classification models' robustness is significantly lower against paraphrases and speech disfluencies, approximately 10\% for both, is surprising, especially given that they attain a higher \texttt{JGA} and \texttt{CorefJGA} by a significant margin. 

On the other hand, it is not surprising that classification models are more robust to unseen entities since their copying mechanism prevents hallucination altogether the distributional shift of correctly predicting spans is smaller than that of copying unseen entities by autoregressively generating from an open vocabulary.\footnote{The advantages of classification models for unseen entities and hallucination needs to be taken with a grain of salt. The copy  mechanism makes TripPy and ConvBERT vulnerable to even simple typos, and to overcome this issue the original implementation defines a mapping for typos that appear in the training set. Since the mapping makes not attempt to directly cover the test set, we keep the same practice, but this unfairly places these models at an advantage over generation models for a test set with similar entities as the training set since the generation models need to memorize typo fixes in order to predict the correct slot.} 
In addition, TripPy models have an advantage for coreference resolution as they observe shorter segments of text since older dialogue history is summarized in the previous dialogue belief state. 
SimpleTOD-style generation models, on the other hand, must examine the entire dialogue history at once. 
Among each class of models, generation models seem to more readily benefit from a pre-finetuning step as both SOLOIST and MUPPET-DST enjoy boosts on most metrics compared to SimpleTOD, which can be considered their baseline. 
However, the large amount of pre-training with open-domain dialogue for ConvBERT-DG seems to be only effective for becoming more robust to speech disfluencies, while other metrics have insignificant differences. 
These results encourage further exploring pre-finetuning generation models. 

\begin{figure}[t!]
    \centering
    \includegraphics[width=\columnwidth]{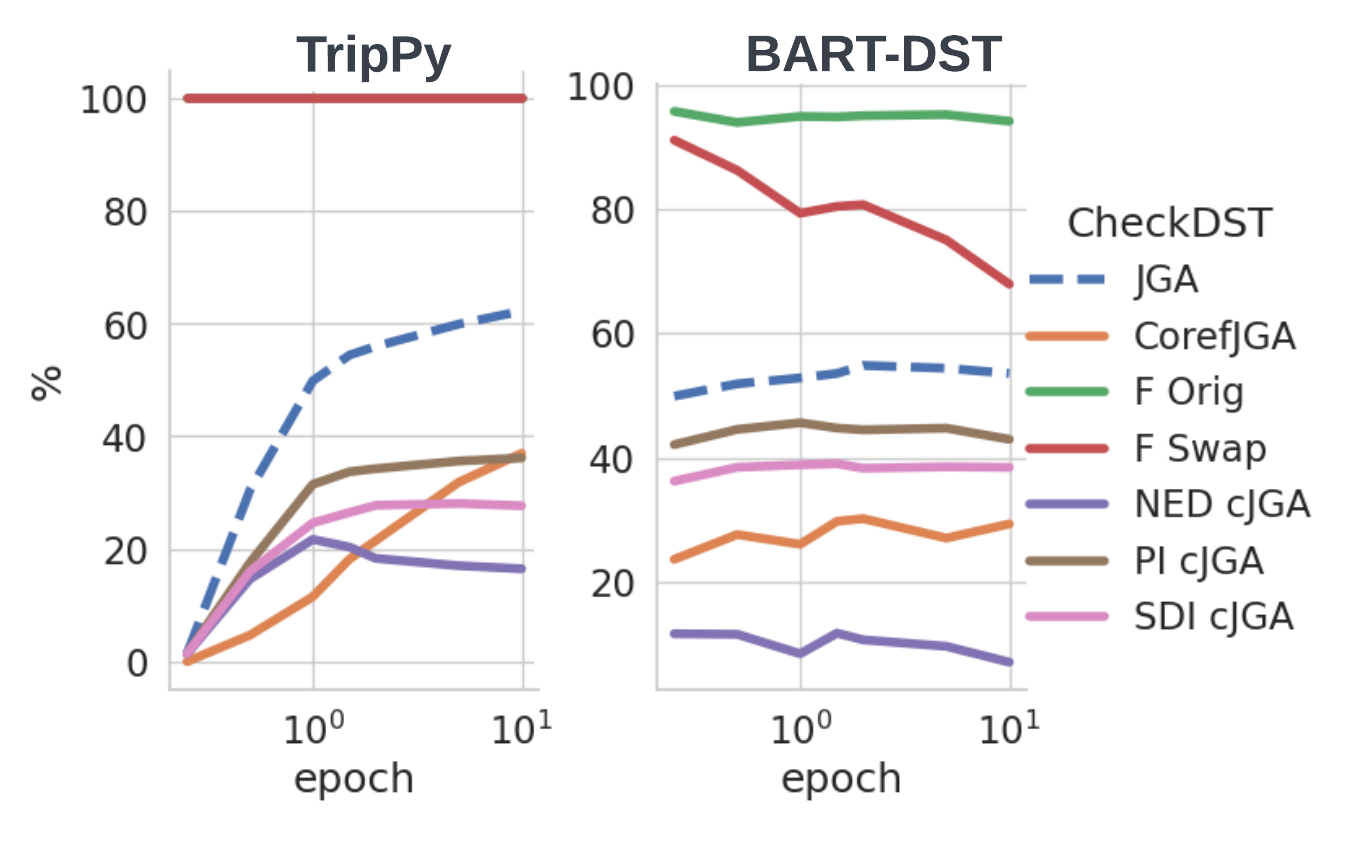}
    \caption{To examine how models overfit in different ways, we plot \ourmetric for intermediate checkpoints. Most of the gains for all metrics except for \texttt{JGA} and \texttt{CorefJGA} are reached before the first few epochs and continue to steadily increase while others stagnate or deteriorate. 
    Classification models show superior performance for perturbations to named entities due to their copy mechanism, but fail to reach similar levels of robustness as generation models for language variety. 
    The $x$-axis uses a logarithmic scale to better visualize the progression in earlier stages of training.
    }
    \vspace{-.15in}
    \label{fig:training_progress}
\end{figure}

\subsection{Intra-model comparison}     
\vspace{-.05in}
The decoupling between \texttt{JGA} and robustness revealed by \ourmetric between in our inter-model comparisons makes us wonder whether a similar problem exists between different checkpoints of the same model. 
We answer this curiosity by running \ourmetric on checkpoints saved every 0.25 epochs until the second epoch and every epoch afterwards to observe how each model's performance on each metric in \ourmetric fares across different checkpoints.
Here, we share our findings from these intra-model comparisons. 

\paragraph{Stopping training early is better for robustness. }
Similar to our findings in inter-model comparisons, we found that even among checkpoints of the same model, higher \texttt{JGA} can lower robustness. 
In \autoref{fig:training_progress}, with TripPy and BART-DST as representative examples, we observe similar trends for both models but with varying degrees. 
While \texttt{JGA} and \texttt{CorefJGA} continue to increase for both models even after the first two epochs, other metrics stagnate or deteriorate, an indication of overfitting.  

To understand the nature of overfitting, we perform qualitative analyses to identify patterns of failure that become apparent over time. 
For each model, we inspect 100 predictions from the NED test set that were correctly predicted by an earlier checkpoint with the highest \texttt{cJGA} and incorrectly predicted by the final checkpoint selected as the best model. 
\vspace{-.1in}

\begin{figure*}[ht!]
    \centering
    \includegraphics[width=\textwidth]{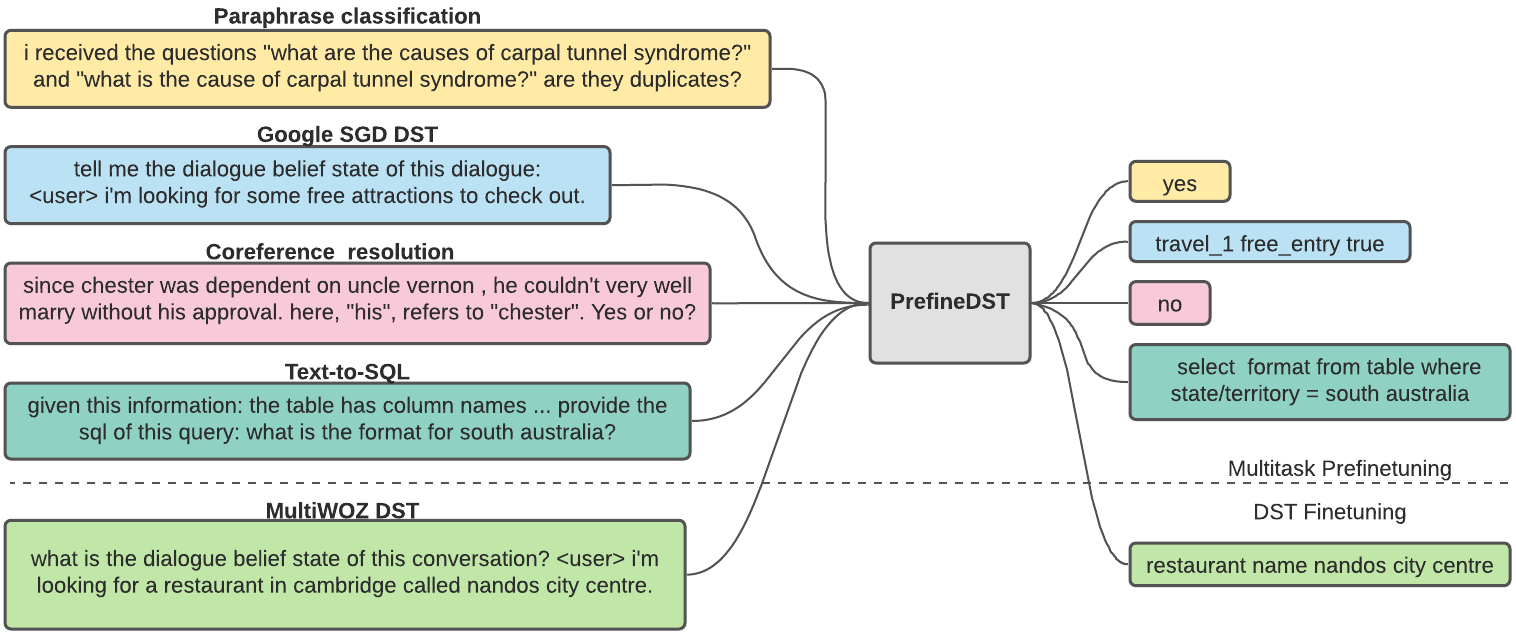}
    \caption{The pre-finetuning step (above dotted line) in \ours is similar to T0 \cite{sanh2021multitask}, using randomly chosen instruction templates to format every task as a generation task. 
    After pre-finetuning, we add a finetuning step for the downstream DST task (below dotted line).
    \justin{can save space here}
    }
    \label{fig:our_model}
\end{figure*}

\paragraph{Classification models give up on span prediction with more training.}
As training progresses, we observe that more than 80\% of these cases for TripPy and ConvBERT-DG become \texttt{none} labels for slot values, indicating that the models are choosing to forgo span predictions over making incorrect span predictions. 
For example, the span for a scrambled entity for the \texttt{restaurant name} slot was correctly predicted to retrieve ``\textit{osdi jkal}'' in the first epoch, reflecting the model's generalizability. 
However, the model with the best \texttt{JGA} predicted that a span for the same slot does not exist, indicating overfitting to entities that they have been repeatedly exposed to during training.

\paragraph{Generation models have difficulty correctly copying out-of-domain slot values.} 
Generation models also struggle with unseen named entities, but their types of failure are more mixed. 
They either (\textit{i}) fail to copy the slot values correctly and produce substrings of the correct slot value or (\textit{ii}) determine that the slot value does not exist and generates nothing, which is equivalent to classification models predicting \texttt{none}. \ab{how often does that happen? your previous plot makes me predict that out of the 80\% drop for NED cJGA, 30\% is attributed to hallucination and 50\% is attributed to generate nothing. is that consistent with your finding or am i hallucinating?}
For the generation models, we saw that about 40\% of the drops are attributed to incorrect predictions while about 60\% is attributed to empty predictions, which is consistent with the drop in $F_{swap}$ and \texttt{NED cJGA} in Figure \ref{fig:training_progress}. 
For instance, in an earlier epoch, BART-DST correctly generates ``\texttt{restaurant name osdi jkal}'', but later instead produces ``\texttt{restaurant name osjkal}''.  
In fewer cases, the prediction becomes empty, similar to the behavior of classification models. 
From this observation, we hypothesize that making the copying process more robust for generation models will significantly boost robustness of generation-based DST models.  

\section{\ours}
\label{sec:ours}
\vspace{-.05in}

The weaknesses exposed by \ourmetric guide us toward approaches that can boost robustness without compromising \texttt{JGA}. 
Inspired by the strong results of massive  multi-task learning on many NLP tasks in recent work, such as MUPPET \cite{aghajanyan2021muppet}, T0 \cite{sanh2021multitask} and FLAN \cite{wei2021finetuned}, and the ease of expanding a generation model to end-to-end task-oriented dialogue \cite{gunasekara2020overview}, we explore a simple multi-tasking approach for a generation model that we name \ours, short for \textit{Pre-finetuned DST}, to train a more well-rounded DST model.

As shown in \autoref{fig:our_model}, \ours first pre-finetunes a pre-trained BART-large model with the same method as T0, but with a narrower set of tasks that can address the robustness issues captured by \ourmetric.
We select tasks that require understanding paraphrases, generating exact spans of text from the context, and resolving coreferences, with the expectation that skills required to perform well on them will be transferred to the fine-tuning step on a downstream DST task and eventually be reflected in better scores on \ourmetric. 
Details on the chosen tasks and implementation details can be found in Appendix \ref{app:prefinedst_details}. 

\paragraph{Pre-finetuning with non-target data is a promising avenue for a robust DST model.}

Overall, results in \autoref{tab:generation_results} show that the simple and intuitive approach behind \ours is successful in maintaining the robustness advantage that generation models have over classification models and performs on-par or better on all \ourmetric metrics among competitive generation model baselines except for on $F_{swap}$. 
This well-rounded performance is maintained even when including classification models, as reflected by the lowest average slack, the difference from the highest scoring model, for all metrics \autoref{tab:avg_distance_from_best}. 

\ours is most directly comparable to SOLOIST and MUPPET in that they all incorporate a pre-finetuning step. 
Thus, it is notable that \ours achieves a higher \texttt{JGA} than both while simultaneously achieving comparable or better results in all dimensions measured by \ourmetric, except for $F_{\texttt{swap}}$. 
This indicates some degree of successful knowledge transfer from the pre-finetuning tasks, but the relatively lower $F_{\texttt{swap}}$ signals that generalizing to correctly copying unseen entities through pre-finetuning is ineffective. 

\justin{qualitative analysis for what kind of knowledge transfer is made when comparing to BART-DST}

\begin{table}[t!]
\begin{adjustbox}{width=\columnwidth,center}
    \centering
    \begingroup
    
    \setlength{\tabcolsep}{3pt} %
    \begin{tabular}{lcccccccccccc} 

        \toprule
        Model & \texttt{JGA} &  \texttt{CorefJGA}& \texttt{PI cJGA} & \texttt{SDI cJGA} & \texttt{NED cJGA} & $F_{\texttt{orig}}$ & $F_{\texttt{swap}}$ \\ \hline

        BART-DST 
        & \colorbox{pink}{$52.5_{0.2}$} & \colorbox{pink}{$25.5_{1.2}$} & \colorbox{pink}{$43.0_{0.8}$} & \colorbox{pink}{$36.5_{1.3}$} & \colorbox{pink}{$9.8_{0.7}$} & $94.8_{0.2}$ & $75.4_{1.8}$ \\ [-0.5em]

        SOLOIST  & $54.8_{0.4}$ & \colorbox{Green}{$30.4_{1.1}$} & \colorbox{pink}{$44.5_{0.6}$} & $38.5_{0.5}$ & $10.7_{0.4}$ & $94.9_{0.1}$ & \colorbox{Green}{$81.1_{1.6}$} \\ [-0.5em]
        MUPPET-DST  & $54.9_{0.4}$ & \colorbox{Green}{$29.9_{1.9}$} & \colorbox{Green}{$45.4_{0.4}$} & $39.1_{0.7}$ & \colorbox{pink}{$7.8_{1.4}$} & $94.6_{0.1}$ & \colorbox{pink}{$68.4_{3.8}$} \\ \hline
        \textbf{PrefineDST} & \colorbox{Green}{$55.7_{0.3}$} & \colorbox{Green}{$30.5_{0.5}$} & \colorbox{Green}{$46.1_{1.0}$} & \colorbox{Green}{$41.8_{1.0}$} & \colorbox{Green}{$11.0_{0.5}$} & $95.2_{0.2}$ & 
        $76.7_{1.4}$ \\
        \bottomrule
    \end{tabular}
    \endgroup
\end{adjustbox}
    \caption{\ourmetric diagnosis results including \ours and other generation models. The annotations are the same as those for \autoref{tab:full_shot_results}. Compared to competitive baselines, \ours improves on the previous highest score for all dimensions except $F_{\texttt{swap}}$.}

    \label{tab:generation_results}
\end{table}

\begin{table}[t!]
\begin{adjustbox}{width=\columnwidth,center}
    \centering
    \tiny
    \begin{tabular}{lr}
    \toprule
        \multicolumn{2}{r}{Average $\Delta$ from Best $\downarrow$} \\ \hline
        TripPy \cite{heck-etal-2020-trippy} & \underline{4.55} \\ [-0.5em]
        ConvBERT-DG \cite{MehriDialoGLUE2020} & 4.57 \\ [-0.5em]
        SimpleTOD \cite{hosseini2020simple} & 6.93 \\ [-0.5em]
        SOLOIST \cite{peng2020soloist} & 4.98 \\ [-0.5em]
        MUPPET-DST \cite{aghajanyan2021muppet} & 5.40 \\    
        \textbf{PrefineDST}& \textbf{3.92} \\ 
        \bottomrule
    \end{tabular}
\end{adjustbox}
    \caption{Average slack of each model from the best performing model on every dimension diagnosed with \ourmetric and \texttt{JGA} based on results in  \autoref{tab:full_shot_results} and \autoref{tab:generation_results}. {\bf Bold} indicates the best performing model and \underline{underline} denotes the second best model. 
    Compared to strong baselines, \ours is the most well-rounded model.
    \justin{define slack in the narrative}
    }    
    \vspace{-.15in}
    \label{tab:avg_distance_from_best}
\end{table}

\ours's superior results to MUPPET-DST, which has been pre-finetuned with more than 40 tasks compared to 8 for \ours, show that choosing NLP tasks that require skill related to the downstream task is more useful than having more tasks. 
Also, the results indicate that multitasking with all tasks as generation tasks is more effective than additional auxiliary layers when DST is also formulated as a generation task. 
In fact, MUPPET's poor performance compared to SimpleTOD on \texttt{NED cJGA} and $F_{swap}$ shows that pre-finetuning can actually be harmful to robustness to unseen entities.  

In conclusion, using a simple and intuitive approach, \ours shows that pre-finetuning with non-target datasets is a promising direction for boosting robustness.
We leave it to future work to leverage \ourmetric as a guide to explore more sophisticated pre-finetuning strategies and non-target tasks to improve on \ours, especially for handling unseen entities.

\section{Related Work}
\label{related_work}

Pretrained language models continue to make impressive strides on NLP benchmarks, surpassing human baseline scores on many of them \cite{lee-etal-2020-squad2, reddy2019coqa, rajpurkar-etal-2016-squad, wang2019superglue, wang-etal-2018-glue}. 
These results lead to questions of whether these models are acquiring the intelligence required for their performance to be robust or instead are taking advantage of spurious correlations \cite{bender-koller-2020-climbing, clark-etal-2019-dont}. Many works show that the latter is the case and seek adversarial techniques to test these models to new limits \cite{gardner-etal-2021-competency, wallace-etal-2019-universal, hosseini2017deceiving} and train them to be more robust \cite{oren-etal-2019-distributionally, jia-etal-2019-certified, jones-etal-2020-robust, zhang-etal-2022-interpreting}. 

Robustness in dialogue models has also been similarly questioned. Perturbations to the dialogue history have exposed that dialogue models do not effectively use dialogue structure information \cite{sankar-etal-2019-neural} and commonsense probes show that they struggle with commonsense reasoning\cite{zhou-etal-2021-probing-commonsense}. 
Specifically for the dialogue state tracking task, several works report drops in performance for conversations with entities unseen during training \cite{qian-etal-2021-annotation, huang2021dair, heck-etal-2020-trippy} or with adversarially created dialogue flows \cite{li2020coco}. 
\newcite{liu-etal-2021-robustness} and \newcite{peng-etal-2021-raddle} recently initiated a rigorous study into the robustness of TOD models to realistic natural language perturbations. 

We extend their work to establish a framework that further facilitates robustness analysis with additional metrics that capture coreference resolution performance and frequency of well-known problems to generation models. Moreover, we propose \texttt{cJGA}, a simple yet rigorous metric that enables measuring robustness in DST without making assumptions about the difficulty of perturbations. 

\ours is motivated by the recent line of work that uses generation models for DST. 
SimpleTOD \cite{hosseini2020simple} first reported viability of formulating TOD tasks in a completely end-to-end manner with a generation model and SOLOIST \cite{peng2020soloist} added a pretraining step to improve on data efficiency. 
\ours, inspired by recent work on impressive results from massive multi-tasking prefinetuning \cite{aghajanyan2021muppet, sanh2021multitask, wei2021finetuned}, extends SimpleTOD and SOLOIST by adding more prefinetuning tasks. 
Concurrent work from \newcite{gupta2022improving} explore a similar procedure for task-oriented dialogue, but their analysis is centered around zero- and few-shot generalization.

\section{Conclusion}

We shared our findings enabled by \ourmetric, a toolkit that standardizes a comprehensive diagnosis of DST models. 
We confirm that benchmark results are insufficient for understanding relative performance as relative robustness is not correlated to benchmark performance. 
We find generation models to be a more promising direction as they are significantly more robust to language variety without making any task-specific design choices.    
We also observed that the standard practice of choosing a checkpoint with validation loss or accuracy is prone to overfitting, exacerbating the trade-off between \texttt{JGA} and robustness for both classification and generation models. 
Finally, we use the robustness issues exposed by \ourmetric to guide the development of \ours, a model that attains well-rounded improvements through a pre-finetuning step that multi-tasks on reasoning skills. 
Moving forward, we encourage future work on task-oriented dialogue to conduct \ourmetric diagnosis for a comprehensive comparison and analysis of DST performance, which we believe will make the search for ideas to build robust task-oriented dialogue models significantly more efficient. 

\section*{Limitations and Broader Impacts}

In this paper, we show that \ourmetric can be used to reveal insights about the robustness of DST models and we hope that the task-oriented dialogue research community will build on and improve \ourmetric as a means for performing holistic comparisons with other work to accelerate our understanding of effective DST approaches. 
We acknowledge that \ourmetric cannot capture generalization to arbitrary distribution shifts in practice as the perturbations against which we measure robustness have to be known ahead of time; and mechanisms to simulate such perturbations need to be built and incorporated, which can be considered a limitation of our work. We also recognize that our analysis has been conducted only in English and therefore our empirical findings may not necessarily be true for DST models built for other languages. 

CheckDST has broad implications as the accurate comparison of robustness and comparative strengths of various DST approaches will help the task-oriented dialogue community take a more informed step towards holistically improving the robustness of best-performing models. 
Consequently, this will improve the quality and accessibility of the numerous services that are powered by task-oriented dialogue models.

\section*{Acknowledgments}

 The work of JM is based in part upon work supported by the Defense Advanced Research Projects Agency (DARPA) under Agreement No. HR00112290025.

\bibliography{anthology,custom}

\begin{thebibliography}{51}
\expandafter\ifx\csname natexlab\endcsname\relax\def\natexlab#1{#1}\fi

\bibitem[{Aghajanyan et~al.(2021)Aghajanyan, Gupta, Shrivastava, Chen,
  Zettlemoyer, and Gupta}]{aghajanyan2021muppet}
Armen Aghajanyan, Anchit Gupta, Akshat Shrivastava, Xilun Chen, Luke
  Zettlemoyer, and Sonal Gupta. 2021.
\newblock Muppet: Massive multi-task representations with pre-finetuning.
\newblock \emph{arXiv preprint arXiv:2101.11038}.

\bibitem[{Bender and Koller(2020)}]{bender-koller-2020-climbing}
Emily~M. Bender and Alexander Koller. 2020.
\newblock \href {https://doi.org/10.18653/v1/2020.acl-main.463} {Climbing
  towards {NLU}: {On} meaning, form, and understanding in the age of data}.
\newblock In \emph{Proceedings of the 58th Annual Meeting of the Association
  for Computational Linguistics}, pages 5185--5198, Online. Association for
  Computational Linguistics.

\bibitem[{Budzianowski et~al.(2018)Budzianowski, Wen, Tseng, Casanueva, Stefan,
  Osman, and Ga{\v{s}}i\'c}]{budzianowski2018large}
Pawe{\l} Budzianowski, Tsung-Hsien Wen, Bo-Hsiang Tseng, I{\~n}igo Casanueva,
  Ultes Stefan, Ramadan Osman, and Milica Ga{\v{s}}i\'c. 2018.
\newblock Multiwoz - a large-scale multi-domain wizard-of-oz dataset for
  task-oriented dialogue modelling.
\newblock In \emph{Proceedings of the 2018 Conference on Empirical Methods in
  Natural Language Processing (EMNLP)}.

\bibitem[{Byrne et~al.(2019)Byrne, Krishnamoorthi, Sankar, Neelakantan,
  Goodrich, Duckworth, Yavuz, Dubey, Kim, and
  Cedilnik}]{byrne-etal-2019-taskmaster}
Bill Byrne, Karthik Krishnamoorthi, Chinnadhurai Sankar, Arvind Neelakantan,
  Ben Goodrich, Daniel Duckworth, Semih Yavuz, Amit Dubey, Kyu-Young Kim, and
  Andy Cedilnik. 2019.
\newblock \href {https://doi.org/10.18653/v1/D19-1459} {Taskmaster-1: Toward a
  realistic and diverse dialog dataset}.
\newblock In \emph{Proceedings of the 2019 Conference on Empirical Methods in
  Natural Language Processing and the 9th International Joint Conference on
  Natural Language Processing (EMNLP-IJCNLP)}, pages 4516--4525, Hong Kong,
  China. Association for Computational Linguistics.

\bibitem[{Chen et~al.(2018)Chen, Zhang, Zhang, and Zhao}]{chen2018quora}
Zihan Chen, Hongbo Zhang, Xiaoji Zhang, and Leqi Zhao. 2018.
\newblock Quora question pairs.
\newblock \emph{URL https://www. kaggle. com/c/quora-question-pairs}.

\bibitem[{Clark et~al.(2019)Clark, Yatskar, and
  Zettlemoyer}]{clark-etal-2019-dont}
Christopher Clark, Mark Yatskar, and Luke Zettlemoyer. 2019.
\newblock \href {https://doi.org/10.18653/v1/D19-1418} {Don{'}t take the easy
  way out: Ensemble based methods for avoiding known dataset biases}.
\newblock In \emph{Proceedings of the 2019 Conference on Empirical Methods in
  Natural Language Processing and the 9th International Joint Conference on
  Natural Language Processing (EMNLP-IJCNLP)}, pages 4069--4082, Hong Kong,
  China. Association for Computational Linguistics.

\bibitem[{Devlin et~al.(2019)Devlin, Chang, Lee, and
  Toutanova}]{devlin-etal-2019-bert}
Jacob Devlin, Ming-Wei Chang, Kenton Lee, and Kristina Toutanova. 2019.
\newblock \href {https://doi.org/10.18653/v1/N19-1423} {{BERT}: Pre-training of
  deep bidirectional transformers for language understanding}.
\newblock In \emph{Proceedings of the 2019 Conference of the North {A}merican
  Chapter of the Association for Computational Linguistics: Human Language
  Technologies, Volume 1 (Long and Short Papers)}, pages 4171--4186,
  Minneapolis, Minnesota. Association for Computational Linguistics.

\bibitem[{Dolan and Brockett(2005)}]{dolan2005msrp}
William~B Dolan and Chris Brockett. 2005.
\newblock Automatically constructing a corpus of sentential paraphrases.
\newblock In \emph{Proceedings of the Third International Workshop on
  Paraphrasing (IWP2005)}.

\bibitem[{Eric et~al.(2020)Eric, Goel, Paul, Sethi, Agarwal, Gao, Kumar, Goyal,
  Ku, and Hakkani-Tur}]{eric2020multiwoz21}
Mihail Eric, Rahul Goel, Shachi Paul, Abhishek Sethi, Sanchit Agarwal, Shuyang
  Gao, Adarsh Kumar, Anuj Goyal, Peter Ku, and Dilek Hakkani-Tur. 2020.
\newblock Multiwoz 2.1: A consolidated multi-domain dialogue dataset with state
  corrections and state tracking baselines.
\newblock In \emph{Proceedings of the 12th Language Resources and Evaluation
  Conference}, pages 422--428.

\bibitem[{Gardner et~al.(2021)Gardner, Merrill, Dodge, Peters, Ross, Singh, and
  Smith}]{gardner-etal-2021-competency}
Matt Gardner, William Merrill, Jesse Dodge, Matthew Peters, Alexis Ross, Sameer
  Singh, and Noah~A. Smith. 2021.
\newblock \href {https://doi.org/10.18653/v1/2021.emnlp-main.135} {Competency
  problems: On finding and removing artifacts in language data}.
\newblock In \emph{Proceedings of the 2021 Conference on Empirical Methods in
  Natural Language Processing}, pages 1801--1813, Online and Punta Cana,
  Dominican Republic. Association for Computational Linguistics.

\bibitem[{Godfrey et~al.(1992)Godfrey, Holliman, and
  McDaniel}]{godfrey1992switchboard}
John~J Godfrey, Edward~C Holliman, and Jane McDaniel. 1992.
\newblock Switchboard: Telephone speech corpus for research and development.
\newblock In \emph{Acoustics, Speech, and Signal Processing, IEEE International
  Conference on}, volume~1, pages 517--520. IEEE Computer Society.

\bibitem[{Gunasekara et~al.(2020)Gunasekara, Kim, D'Haro, Rastogi, Chen, Eric,
  Hedayatnia, Gopalakrishnan, Liu, Huang et~al.}]{gunasekara2020overview}
Chulaka Gunasekara, Seokhwan Kim, Luis~Fernando D'Haro, Abhinav Rastogi,
  Yun-Nung Chen, Mihail Eric, Behnam Hedayatnia, Karthik Gopalakrishnan, Yang
  Liu, Chao-Wei Huang, et~al. 2020.
\newblock Overview of the ninth dialog system technology challenge: Dstc9.
\newblock \emph{arXiv preprint arXiv:2011.06486}.

\bibitem[{Gupta et~al.(2022)Gupta, Jiao, Yeh, Mehri, Eskenazi, and
  Bigham}]{gupta2022improving}
Prakhar Gupta, Cathy Jiao, Yi-Ting Yeh, Shikib Mehri, Maxine Eskenazi, and
  Jeffrey~P Bigham. 2022.
\newblock Improving zero and few-shot generalization in dialogue through
  instruction tuning.
\newblock \emph{arXiv preprint arXiv:2205.12673}.

\bibitem[{Han et~al.(2020)Han, Liu, Takanobu, Lian, Huang, Wan, Peng, and
  Huang}]{han2020multiwoz23}
Ting Han, Ximing Liu, Ryuichi Takanobu, Yixin Lian, Chongxuan Huang, Dazhen
  Wan, Wei Peng, and Minlie Huang. 2020.
\newblock Multiwoz 2.3: A multi-domain task-oriented dialogue dataset enhanced
  with annotation corrections and co-reference annotation.
\newblock \emph{arXiv preprint arXiv:2010.05594}.

\bibitem[{Heck et~al.(2020)Heck, van Niekerk, Lubis, Geishauser, Lin, Moresi,
  and Gasic}]{heck-etal-2020-trippy}
Michael Heck, Carel van Niekerk, Nurul Lubis, Christian Geishauser, Hsien-Chin
  Lin, Marco Moresi, and Milica Gasic. 2020.
\newblock \href {https://aclanthology.org/2020.sigdial-1.4} {{T}rip{P}y: A
  triple copy strategy for value independent neural dialog state tracking}.
\newblock In \emph{Proceedings of the 21th Annual Meeting of the Special
  Interest Group on Discourse and Dialogue}, pages 35--44, 1st virtual meeting.
  Association for Computational Linguistics.

\bibitem[{Hosseini et~al.(2017)Hosseini, Kannan, Zhang, and
  Poovendran}]{hosseini2017deceiving}
Hossein Hosseini, Sreeram Kannan, Baosen Zhang, and Radha Poovendran. 2017.
\newblock Deceiving google's perspective api built for detecting toxic
  comments.
\newblock \emph{arXiv preprint arXiv:1702.08138}.

\bibitem[{Hosseini-Asl et~al.(2020)Hosseini-Asl, McCann, Wu, Yavuz, and
  Socher}]{hosseini2020simple}
Ehsan Hosseini-Asl, Bryan McCann, Chien-Sheng Wu, Semih Yavuz, and Richard
  Socher. 2020.
\newblock A simple language model for task-oriented dialogue.
\newblock \emph{arXiv preprint arXiv:2005.00796}.

\bibitem[{Huang et~al.(2021)Huang, Halbe, Sankar, Amini, Kottur, Geramifard,
  Razaviyayn, and Beirami}]{huang2021dair}
Tianjian Huang, Shaunak Halbe, Chinnadhurai Sankar, Pooyan Amini, Satwik
  Kottur, Alborz Geramifard, Meisam Razaviyayn, and Ahmad Beirami. 2021.
\newblock \href {http://arxiv.org/abs/2110.11205} {Dair: Data augmented
  invariant regularization}.

\bibitem[{Jia et~al.(2019)Jia, Raghunathan, G{\"o}ksel, and
  Liang}]{jia-etal-2019-certified}
Robin Jia, Aditi Raghunathan, Kerem G{\"o}ksel, and Percy Liang. 2019.
\newblock \href {https://doi.org/10.18653/v1/D19-1423} {Certified robustness to
  adversarial word substitutions}.
\newblock In \emph{Proceedings of the 2019 Conference on Empirical Methods in
  Natural Language Processing and the 9th International Joint Conference on
  Natural Language Processing (EMNLP-IJCNLP)}, pages 4129--4142, Hong Kong,
  China. Association for Computational Linguistics.

\bibitem[{Jones et~al.(2020)Jones, Jia, Raghunathan, and
  Liang}]{jones-etal-2020-robust}
Erik Jones, Robin Jia, Aditi Raghunathan, and Percy Liang. 2020.
\newblock \href {https://doi.org/10.18653/v1/2020.acl-main.245} {Robust
  encodings: A framework for combating adversarial typos}.
\newblock In \emph{Proceedings of the 58th Annual Meeting of the Association
  for Computational Linguistics}, pages 2752--2765, Online. Association for
  Computational Linguistics.

\bibitem[{Lee et~al.(2020)Lee, Hwang, and Cho}]{lee-etal-2020-squad2}
Gyeongbok Lee, Seung-won Hwang, and Hyunsouk Cho. 2020.
\newblock \href {https://aclanthology.org/2020.lrec-1.667} {{SQ}u{AD}2-{CR}:
  Semi-supervised annotation for cause and rationales for unanswerability in
  {SQ}u{AD} 2.0}.
\newblock In \emph{Proceedings of the 12th Language Resources and Evaluation
  Conference}, pages 5425--5432, Marseille, France. European Language Resources
  Association.

\bibitem[{Levesque et~al.(2012)Levesque, Davis, and
  Morgenstern}]{levesque2012winograd}
Hector Levesque, Ernest Davis, and Leora Morgenstern. 2012.
\newblock The winograd schema challenge.
\newblock In \emph{Thirteenth International Conference on the Principles of
  Knowledge Representation and Reasoning}.

\bibitem[{Lewis et~al.(2020)Lewis, Liu, Goyal, Ghazvininejad, Mohamed, Levy,
  Stoyanov, and Zettlemoyer}]{lewis-etal-2020-bart}
Mike Lewis, Yinhan Liu, Naman Goyal, Marjan Ghazvininejad, Abdelrahman Mohamed,
  Omer Levy, Veselin Stoyanov, and Luke Zettlemoyer. 2020.
\newblock \href {https://doi.org/10.18653/v1/2020.acl-main.703} {{BART}:
  Denoising sequence-to-sequence pre-training for natural language generation,
  translation, and comprehension}.
\newblock In \emph{Proceedings of the 58th Annual Meeting of the Association
  for Computational Linguistics}, pages 7871--7880, Online. Association for
  Computational Linguistics.

\bibitem[{Li et~al.(2020)Li, Yavuz, Hashimoto, Li, Niu, Rajani, Yan, Zhou, and
  Xiong}]{li2020coco}
Shiyang Li, Semih Yavuz, Kazuma Hashimoto, Jia Li, Tong Niu, Nazneen Rajani,
  Xifeng Yan, Yingbo Zhou, and Caiming Xiong. 2020.
\newblock Coco: Controllable counterfactuals for evaluating dialogue state
  trackers.
\newblock In \emph{International Conference on Learning Representations}.

\bibitem[{Liu et~al.(2021)Liu, Takanobu, Wen, Wan, Li, Nie, Li, Peng, and
  Huang}]{liu-etal-2021-robustness}
Jiexi Liu, Ryuichi Takanobu, Jiaxin Wen, Dazhen Wan, Hongguang Li, Weiran Nie,
  Cheng Li, Wei Peng, and Minlie Huang. 2021.
\newblock \href {https://doi.org/10.18653/v1/2021.acl-long.192} {Robustness
  testing of language understanding in task-oriented dialog}.
\newblock In \emph{Proceedings of the 59th Annual Meeting of the Association
  for Computational Linguistics and the 11th International Joint Conference on
  Natural Language Processing (Volume 1: Long Papers)}, pages 2467--2480,
  Online. Association for Computational Linguistics.

\bibitem[{Massarelli et~al.(2020)Massarelli, Petroni, Piktus, Ott,
  Rockt{\"a}schel, Plachouras, Silvestri, and
  Riedel}]{massarelli-etal-2020-decoding}
Luca Massarelli, Fabio Petroni, Aleksandra Piktus, Myle Ott, Tim
  Rockt{\"a}schel, Vassilis Plachouras, Fabrizio Silvestri, and Sebastian
  Riedel. 2020.
\newblock \href {https://doi.org/10.18653/v1/2020.findings-emnlp.22} {How
  decoding strategies affect the verifiability of generated text}.
\newblock In \emph{Findings of the Association for Computational Linguistics:
  EMNLP 2020}, pages 223--235, Online. Association for Computational
  Linguistics.

\bibitem[{Maynez et~al.(2020)Maynez, Narayan, Bohnet, and
  McDonald}]{maynez-etal-2020-faithfulness}
Joshua Maynez, Shashi Narayan, Bernd Bohnet, and Ryan McDonald. 2020.
\newblock \href {https://doi.org/10.18653/v1/2020.acl-main.173} {On
  faithfulness and factuality in abstractive summarization}.
\newblock In \emph{Proceedings of the 58th Annual Meeting of the Association
  for Computational Linguistics}, pages 1906--1919, Online. Association for
  Computational Linguistics.

\bibitem[{Mehri et~al.(2020)Mehri, Eric, and Hakkani-Tur}]{MehriDialoGLUE2020}
Shikib Mehri, Mihail Eric, and Dilek Hakkani-Tur. 2020.
\newblock {DialoGLUE}: A natural language understanding benchmark for
  task-oriented dialogue.
\newblock \emph{ArXiv}, abs/2009.13570.

\bibitem[{Miller et~al.(2017)Miller, Feng, Batra, Bordes, Fisch, Lu, Parikh,
  and Weston}]{miller-etal-2017-parlai}
Alexander Miller, Will Feng, Dhruv Batra, Antoine Bordes, Adam Fisch, Jiasen
  Lu, Devi Parikh, and Jason Weston. 2017.
\newblock \href {https://doi.org/10.18653/v1/D17-2014} {{P}arl{AI}: A dialog
  research software platform}.
\newblock In \emph{Proceedings of the 2017 Conference on Empirical Methods in
  Natural Language Processing: System Demonstrations}, pages 79--84,
  Copenhagen, Denmark. Association for Computational Linguistics.

\bibitem[{Oren et~al.(2019)Oren, Sagawa, Hashimoto, and
  Liang}]{oren-etal-2019-distributionally}
Yonatan Oren, Shiori Sagawa, Tatsunori~B. Hashimoto, and Percy Liang. 2019.
\newblock \href {https://doi.org/10.18653/v1/D19-1432} {Distributionally robust
  language modeling}.
\newblock In \emph{Proceedings of the 2019 Conference on Empirical Methods in
  Natural Language Processing and the 9th International Joint Conference on
  Natural Language Processing (EMNLP-IJCNLP)}, pages 4227--4237, Hong Kong,
  China. Association for Computational Linguistics.

\bibitem[{Peng et~al.(2021{\natexlab{a}})Peng, Li, Li, Shayandeh, Liden, and
  Gao}]{peng2020soloist}
Baolin Peng, Chunyuan Li, Jinchao Li, Shahin Shayandeh, Lars Liden, and
  Jianfeng Gao. 2021{\natexlab{a}}.
\newblock \href {https://doi.org/10.1162/tacl_a_00399} {Soloist: Building task
  bots at scale with transfer learning and machine teaching}.
\newblock \emph{Transactions of the Association for Computational Linguistics},
  9:807--824.

\bibitem[{Peng et~al.(2021{\natexlab{b}})Peng, Li, Zhang, Zhu, Li, and
  Gao}]{peng-etal-2021-raddle}
Baolin Peng, Chunyuan Li, Zhu Zhang, Chenguang Zhu, Jinchao Li, and Jianfeng
  Gao. 2021{\natexlab{b}}.
\newblock \href {https://doi.org/10.18653/v1/2021.acl-long.341} {{RADDLE}: An
  evaluation benchmark and analysis platform for robust task-oriented dialog
  systems}.
\newblock In \emph{Proceedings of the 59th Annual Meeting of the Association
  for Computational Linguistics and the 11th International Joint Conference on
  Natural Language Processing (Volume 1: Long Papers)}, pages 4418--4429,
  Online. Association for Computational Linguistics.

\bibitem[{Peng et~al.(2020)Peng, Zhu, Li, Li, Li, Zeng, and
  Gao}]{peng-etal-2020-shot}
Baolin Peng, Chenguang Zhu, Chunyuan Li, Xiujun Li, Jinchao Li, Michael Zeng,
  and Jianfeng Gao. 2020.
\newblock \href {https://doi.org/10.18653/v1/2020.findings-emnlp.17} {Few-shot
  natural language generation for task-oriented dialog}.
\newblock In \emph{Findings of the Association for Computational Linguistics:
  EMNLP 2020}, pages 172--182, Online. Association for Computational
  Linguistics.

\bibitem[{Qian et~al.(2021)Qian, Beirami, Lin, De, Geramifard, Yu, and
  Sankar}]{qian-etal-2021-annotation}
Kun Qian, Ahmad Beirami, Zhouhan Lin, Ankita De, Alborz Geramifard, Zhou Yu,
  and Chinnadhurai Sankar. 2021.
\newblock \href {https://aclanthology.org/2021.sigdial-1.35} {Annotation
  inconsistency and entity bias in {M}ulti{WOZ}}.
\newblock In \emph{Proceedings of the 22nd Annual Meeting of the Special
  Interest Group on Discourse and Dialogue}, pages 326--337, Singapore and
  Online. Association for Computational Linguistics.

\bibitem[{Quan et~al.(2019)Quan, Xiong, Webber, and Hu}]{quan-etal-2019-gecor}
Jun Quan, Deyi Xiong, Bonnie Webber, and Changjian Hu. 2019.
\newblock \href {https://doi.org/10.18653/v1/D19-1462} {{GECOR}: An end-to-end
  generative ellipsis and co-reference resolution model for task-oriented
  dialogue}.
\newblock In \emph{Proceedings of the 2019 Conference on Empirical Methods in
  Natural Language Processing and the 9th International Joint Conference on
  Natural Language Processing (EMNLP-IJCNLP)}, pages 4547--4557, Hong Kong,
  China. Association for Computational Linguistics.

\bibitem[{Rajpurkar et~al.(2018)Rajpurkar, Jia, and
  Liang}]{rajpurkar-etal-2018-squadv2}
Pranav Rajpurkar, Robin Jia, and Percy Liang. 2018.
\newblock \href {https://doi.org/10.18653/v1/P18-2124} {Know what you don{'}t
  know: Unanswerable questions for {SQ}u{AD}}.
\newblock In \emph{Proceedings of the 56th Annual Meeting of the Association
  for Computational Linguistics (Volume 2: Short Papers)}, pages 784--789,
  Melbourne, Australia. Association for Computational Linguistics.

\bibitem[{Rajpurkar et~al.(2016)Rajpurkar, Zhang, Lopyrev, and
  Liang}]{rajpurkar-etal-2016-squad}
Pranav Rajpurkar, Jian Zhang, Konstantin Lopyrev, and Percy Liang. 2016.
\newblock \href {https://doi.org/10.18653/v1/D16-1264} {{SQ}u{AD}: 100,000+
  questions for machine comprehension of text}.
\newblock In \emph{Proceedings of the 2016 Conference on Empirical Methods in
  Natural Language Processing}, pages 2383--2392, Austin, Texas. Association
  for Computational Linguistics.

\bibitem[{Rastogi et~al.(2020)Rastogi, Zang, Sunkara, Gupta, and
  Khaitan}]{rastogi2020towardssgd}
Abhinav Rastogi, Xiaoxue Zang, Srinivas Sunkara, Raghav Gupta, and Pranav
  Khaitan. 2020.
\newblock Towards scalable multi-domain conversational agents: The
  schema-guided dialogue dataset.
\newblock In \emph{Proceedings of the AAAI Conference on Artificial
  Intelligence}, volume~34, pages 8689--8696.

\bibitem[{Reddy et~al.(2019)Reddy, Chen, and Manning}]{reddy2019coqa}
Siva Reddy, Danqi Chen, and Christopher~D Manning. 2019.
\newblock Coqa: A conversational question answering challenge.
\newblock \emph{Transactions of the Association for Computational Linguistics},
  7:249--266.

\bibitem[{Ribeiro et~al.(2020)Ribeiro, Wu, Guestrin, and
  Singh}]{ribeiro-etal-2020-beyond}
Marco~Tulio Ribeiro, Tongshuang Wu, Carlos Guestrin, and Sameer Singh. 2020.
\newblock \href {https://doi.org/10.18653/v1/2020.acl-main.442} {Beyond
  accuracy: Behavioral testing of {NLP} models with {C}heck{L}ist}.
\newblock In \emph{Proceedings of the 58th Annual Meeting of the Association
  for Computational Linguistics}, pages 4902--4912, Online. Association for
  Computational Linguistics.

\bibitem[{Sanh et~al.(2021)Sanh, Webson, Raffel, Bach, Sutawika, Alyafeai,
  Chaffin, Stiegler, Scao, Raja et~al.}]{sanh2021multitask}
Victor Sanh, Albert Webson, Colin Raffel, Stephen~H Bach, Lintang Sutawika,
  Zaid Alyafeai, Antoine Chaffin, Arnaud Stiegler, Teven~Le Scao, Arun Raja,
  et~al. 2021.
\newblock Multitask prompted training enables zero-shot task generalization.
\newblock \emph{arXiv preprint arXiv:2110.08207}.

\bibitem[{Sankar et~al.(2019)Sankar, Subramanian, Pal, Chandar, and
  Bengio}]{sankar-etal-2019-neural}
Chinnadhurai Sankar, Sandeep Subramanian, Chris Pal, Sarath Chandar, and Yoshua
  Bengio. 2019.
\newblock \href {https://doi.org/10.18653/v1/P19-1004} {Do neural dialog
  systems use the conversation history effectively? an empirical study}.
\newblock In \emph{Proceedings of the 57th Annual Meeting of the Association
  for Computational Linguistics}, pages 32--37, Florence, Italy. Association
  for Computational Linguistics.

\bibitem[{Su et~al.(2021)Su, Shu, Mansimov, Gupta, Cai, Lai, and
  Zhang}]{su2021pptod}
Yixuan Su, Lei Shu, Elman Mansimov, Arshit Gupta, Deng Cai, Yi{-}An Lai, and
  Yi~Zhang. 2021.
\newblock \href {http://arxiv.org/abs/2109.14739} {Multi-task pre-training for
  plug-and-play task-oriented dialogue system}.
\newblock \emph{CoRR}, abs/2109.14739.

\bibitem[{Wallace et~al.(2019)Wallace, Feng, Kandpal, Gardner, and
  Singh}]{wallace-etal-2019-universal}
Eric Wallace, Shi Feng, Nikhil Kandpal, Matt Gardner, and Sameer Singh. 2019.
\newblock \href {https://doi.org/10.18653/v1/D19-1221} {Universal adversarial
  triggers for attacking and analyzing {NLP}}.
\newblock In \emph{Proceedings of the 2019 Conference on Empirical Methods in
  Natural Language Processing and the 9th International Joint Conference on
  Natural Language Processing (EMNLP-IJCNLP)}, pages 2153--2162, Hong Kong,
  China. Association for Computational Linguistics.

\bibitem[{Wang et~al.(2019)Wang, Pruksachatkun, Nangia, Singh, Michael, Hill,
  Levy, and Bowman}]{wang2019superglue}
Alex Wang, Yada Pruksachatkun, Nikita Nangia, Amanpreet Singh, Julian Michael,
  Felix Hill, Omer Levy, and Samuel~R Bowman. 2019.
\newblock Superglue: a stickier benchmark for general-purpose language
  understanding systems.
\newblock In \emph{Proceedings of the 33rd International Conference on Neural
  Information Processing Systems}, pages 3266--3280.

\bibitem[{Wang et~al.(2018)Wang, Singh, Michael, Hill, Levy, and
  Bowman}]{wang-etal-2018-glue}
Alex Wang, Amanpreet Singh, Julian Michael, Felix Hill, Omer Levy, and Samuel
  Bowman. 2018.
\newblock \href {https://doi.org/10.18653/v1/W18-5446} {{GLUE}: A multi-task
  benchmark and analysis platform for natural language understanding}.
\newblock In \emph{Proceedings of the 2018 {EMNLP} Workshop {B}lackbox{NLP}:
  Analyzing and Interpreting Neural Networks for {NLP}}, pages 353--355,
  Brussels, Belgium. Association for Computational Linguistics.

\bibitem[{Wang et~al.(2020)Wang, Che, Liu, Qin, Liu, and Wang}]{wang2020multi}
Shaolei Wang, Wangxiang Che, Qi~Liu, Pengda Qin, Ting Liu, and William~Yang
  Wang. 2020.
\newblock Multi-task self-supervised learning for disfluency detection.
\newblock In \emph{Proceedings of the AAAI Conference on Artificial
  Intelligence}, volume~34, pages 9193--9200.

\bibitem[{Wei et~al.(2021)Wei, Bosma, Zhao, Guu, Yu, Lester, Du, Dai, and
  Le}]{wei2021finetuned}
Jason Wei, Maarten Bosma, Vincent~Y Zhao, Kelvin Guu, Adams~Wei Yu, Brian
  Lester, Nan Du, Andrew~M Dai, and Quoc~V Le. 2021.
\newblock Finetuned language models are zero-shot learners.
\newblock \emph{arXiv preprint arXiv:2109.01652}.

\bibitem[{Zhang et~al.(2022)Zhang, Pan, Tan, and
  Kan}]{zhang-etal-2022-interpreting}
Yunxiang Zhang, Liangming Pan, Samson Tan, and Min-Yen Kan. 2022.
\newblock \href {https://doi.org/10.18653/v1/2022.findings-acl.315}
  {Interpreting the robustness of neural {NLP} models to textual
  perturbations}.
\newblock In \emph{Findings of the Association for Computational Linguistics:
  ACL 2022}, pages 3993--4007, Dublin, Ireland. Association for Computational
  Linguistics.

\bibitem[{Zhong et~al.(2017)Zhong, Xiong, and Socher}]{zhong2017seq2sql}
Victor Zhong, Caiming Xiong, and Richard Socher. 2017.
\newblock Seq2sql: Generating structured queries from natural language using
  reinforcement learning.
\newblock \emph{arXiv preprint arXiv:1709.00103}.

\bibitem[{Zhou et~al.(2021)Zhou, Jandaghi, Cho, Lin, Pujara, and
  Ren}]{zhou-etal-2021-probing-commonsense}
Pei Zhou, Pegah Jandaghi, Hyundong Cho, Bill~Yuchen Lin, Jay Pujara, and Xiang
  Ren. 2021.
\newblock \href {https://doi.org/10.18653/v1/2021.findings-emnlp.349} {Probing
  commonsense explanation in dialogue response generation}.
\newblock In \emph{Findings of the Association for Computational Linguistics:
  EMNLP 2021}, pages 4132--4146, Punta Cana, Dominican Republic. Association
  for Computational Linguistics.

\end{thebibliography}
\bibliographystyle{acl_natbib}

\appendix

\clearpage
\section*{Appendix}
\label{sec:appendix}

\section{Further Justification for \texttt{cJGA}}
\label{app:cJGA}

\begin{lemma}
Let 
\begin{align}
    \texttt{JGA} & := \frac{1}{n} \sum_{i \in [n]} f(z_i ; \theta),\\
    \widetilde{\texttt{JGA}} & := \frac{1}{n} \sum_{i \in [n]} f(\widetilde{z_i} ; \theta), \\
        \texttt{cJGA} &:= \frac{1}{n} \sum_{i \in [n]} \mathbf{1}(f(z_i; \theta) = f(\widetilde{z_i}; \theta) = 1),
    \label{eq:def-cjga}
\end{align}
where $\mathbf{1}(\cdot)$ denotes the indicator function.
Then,
\begin{equation}
    \texttt{cJGA}  \leq \min\{ \texttt{JGA} , \widetilde{\texttt{JGA}} , 1 - |\texttt{JGA} - \widetilde{\texttt{JGA}}|\}.
\end{equation}
\label{lem:cJGA}
\end{lemma}
\begin{proof}
It is clear that $\texttt{cJGA}  \leq \min\{ \texttt{JGA} , \widetilde{\texttt{JGA}} \}$. 
The proof to show that $\texttt{cJGA} \leq 1 - |\texttt{JGA} - \widetilde{\texttt{JGA}}|$ follows from the following set of inequalities:
\begin{align}
   \texttt{cJGA} &= \frac{1}{n}\sum_{i \in [n]} \mathbf{1}(f(z_i; \theta) = f(\widetilde{z_i}; \theta) = 1)\nonumber\\ 
   &\leq  1 - \frac{1}{n}\sum_{i \in [n]}|f(z_i; \theta) - f(\widetilde{z_i}; \theta)| \label{eq:0} \\
   & \leq 1 - \left|\frac{1}{n}\sum_{i \in [n]} (f(z_i; \theta) - f(\widetilde{z_i}; \theta))\right|\label{eq:1}\\
    & = 1 -  |\texttt{JGA} - \widetilde{\texttt{JGA}}|,\label{eq:3}\nonumber
\end{align}
where \eqref{eq:0} follows from the fact that for any $i \in [n],$
\begin{align*}
\resizebox{\linewidth}{!}{
   $\mathbf{1}(f(z_i; \theta) = f(\widetilde{z_i}; \theta) = 1) \leq   1 - |f(z_i; \theta) - f(\widetilde{z_i}; \theta)|$}
\end{align*}
and \eqref{eq:1} follows from Jensen's inequality.
\end{proof}
Lemma~\ref{lem:cJGA} shows that \texttt{cJGA} not only captures the discrepancy between \texttt{JGA} and $\widetilde{\texttt{JGA}}$, but it can actually capture counterfactual robustness beyond that.
As an example, consider a case where $\texttt{JGA} = \widetilde{\texttt{JGA}} = 0.6$, hence no drop is observed. In this case if $\texttt{cJGA} \approx 0.6,$ it means that the performance is robust but the model is struggling with learning some particular flows. On the other hand, if \texttt{cJGA} is low, e.g., 0.2, it means that the performance is statistically fragile and the \texttt{JGA} is mostly affected by model robustness. This would not have been revealed by solely quantifying the \texttt{JGA} drop. As a second example, consider a case where the $\texttt{JGA} = 0.8$ whereas $\widetilde{\texttt{JGA}} = 0.6$. It is straightforward to show that \texttt{cJGA} cannot be larger than $0.75$ (see Lemma~\ref{lem:cJGA}), hence capturing the \texttt{JGA} drop. On the other hand, \texttt{cJGA} may be (much) smaller than $0.75$ if there are further statistical model variations due to lack of robustness (inconsistency of performance across original and perturbed samples), which would not be revealed by the \texttt{JGA} drop.

\section{Further notes on \ourmetric}
\label{app:checkdst_appendix}

\subsection{Results on MultiWOZ2.1}

Our preliminary experiments with MultiWOZ2.1~\cite{eric2020multiwoz21} showed similar results as MultiWOZ2.3. 
We shifted to using the latter because it has cleaner annotations that enable more accurate follow-up diagnosis.
However, acknowledging that many results are still provided using 2.1, we report PrefineDST's JGA on MultiWOZ2.1, which is $53.8 \pm 0.3$. 
BART-DST's JGA is $51.5 \pm 0.2$. 
There is a relatively smaller gap than that for MultiWOZ2.3, which is in line with results for other models shown in the official MultiWOZ2.3 benchmark.\footnote{\url{https://github.com/lexmen318/MultiWOZ-coref}} 

\subsection{Generalizability of CheckDST}
\label{app:checkdst_generalizability}

For CheckDST to be applied to a TOD dataset, the dataset must have dialogue act and belief state annotations at the minimum. 
If these annotations are available, the LAUG toolkit can be used to insert speech disfluencies and generate paraphrases with a SC-GPT model \cite{peng-etal-2020-shot, liu-etal-2021-robustness}. 
To replace named entities, named entity slot types must be pre-defined such that these values can be automatically scrambled or replaced, both in the annotations and dialogue. 
In the same vein, the named entity slot types are used to determine hallucination frequency by measuring how often their slot values are not values from the given text. 
\texttt{CorefJGA} is the least portable metric in CheckDST as it requires coreference annotations. 
However, using simple regular expressions for pronouns and frequently used terms such as \textit{``same $X$ as''} can discover many coreference cases with high precision.
These subsets can then be used for measuring \texttt{CorefJGA}.

\subsection{Baseline Training Details}
\label{app:baseline_training}

Most models are trained on MultiWOZ~2.1 \cite{eric2020multiwoz21} and therefore we retrain them on MultiWOZ~2.3 \cite{han2020multiwoz23} before assessing them on \ourmetric. 
Unless otherwise specified, we use the set of hyperparameters mentioned by the original work and run five iterations with different seed values for results to have more statistical significance. 
If not provided, we do a hyperparameter search for the best learning rate and choose the configuration that leads to the best median \texttt{JGA} on the validation set. 
For each baseline, we train with five different seeds and report the median and standard error of these runs.

For finetuning MUPPET \cite{aghajanyan2021muppet} with MultiWOZ, we follow the same setup used in the original work for finetuning on downstream tasks. 
We drop the additional layers and use only the parameters that are part of the original BART architecture to finetune MUPPET on MultiWOZ in the same way as BART-DST. 

For intra-model comparisons, we conduct \ourmetric diagnosis on checkpoints at the following epochs: [0.25, 0.5, 0.75, 1, 1.5, 2, 5, 10].

\begin{table*}[h]
\begin{adjustbox}{width=\textwidth,center}
    \centering
    \begin{tabular}{llrr} 
        Dataset & Type & Train / Valid / Test Size & Targeted \ourmetric metrics \\ \hline
        MSR \cite{dolan2005msrp} & Paraphrase & 4,076 / 862 / 863 & PI cJGA \\ [-0.5em]
        QQP \cite{chen2018quora} & Paraphrase & 305,408 / 38,176 / 38,176 & PI cJGA \\ [-0.5em] 
        WSC* \cite{levesque2012winograd} & Coref & 554 / 104 & CorefJGA \\ [-0.5em]
        WNLI* \cite{ wang-etal-2018-glue} & Coref & 635 / 71 & CorefJGA \\ [-0.5em]
        SQuAD v2*  \cite{rajpurkar-etal-2018-squadv2} & Q\&A & 130,319 / 11,873 & NEI cJGA, NoHF  \\ [-0.5em]
        CoQA*  \cite{reddy2019coqa} & Q\&A & 108,647 / 7,983 & NEI cJGA, NoHF, CorefJGA \\[-0.5em]
        WikiSQL \cite{zhong2017seq2sql} & Text-SQL & 56,355 / 8,421 / 15,878& NEI cJGA, NoHF \\ [-0.5em]
        SGD \cite{rastogi2020towardssgd} & TOD & 164,982 / 24,363 / 42,297 & NEI cJGA, NoHF, CorefJGA
    \end{tabular}
\end{adjustbox}
    \caption{A summary of prefinetuning datasets that we use for \ours. *These datasets do not have a separate test set. We reuse the validation set for these datasets.}
    \label{tab:prefinetuning_tasks}
\end{table*}

\section{\ours Details}
\label{app:prefinedst_details}

\subsection{Implementation details}

\paragraph{Task formulation.}
We take the same approach as T0 in uniformly formatting all datasets, reusing prompts for tasks that are already used for T0 and designing new ones for those that are not. For each example from a dataset, we randomly sample from a corresponding set of instruction templates and modify each sample according to the chosen template. 

\paragraph{Prompts.}

For tasks that are not used in T0 such as WikiSQL \cite{zhong2017seq2sql} and SGD \cite{rastogi2020towardssgd}, we modify applicable prompts from different tasks to create at least five different prompt templates for each task.
One of these templates are randomly chosen for training time and inference time. The random seed is changed during training time but kept the same at test time to ensure replicability. 

\paragraph{Training details.}
Following \newcite{sanh2021multitask}, we do not adjust the sampling rate based on the sample size of each task that we multitask with during prefinetuning. 
Since all tasks are formatted as a sequence-to-sequence generation task, we do not need any additional layers as was needed for MUPPET nor form heterogeneous batches that contain samples from multiple tasks. 
For the prefinetuning step, we do a hyperparameter search with only five different learning rates and keep the batch size at 64 per GPU to find the model with the lowest loss value on the test set. We use 8 A100 GPUs and train for 10 epochs, early stopping on the loss value of the validation set with a patience of 3. This process amounts to a total of approximately 400 GPU hours. We get best results with a learning rate of $1e^{-5}$. 

Then, we finetune the prefinetuned model. We vary both the learning rate and the batch size and train for 10 epochs on a single A100 GPU, running five iterations with different seed values, after which we choose the checkpoint with the best \texttt{JGA} on the validation set. The best performing model uses a batch size of 4 and learning rate of $5e^{-5}$. This amounts to about 170 GPU hours in total. 
We use ParlAI \cite{miller-etal-2017-parlai} for all of our experiments. 

\subsection{Prefinetuning Tasks}

We choose prefinetuning tasks based on their intuitive potential for improving on qualities measured by \ourmetric. 
They can largely be categorized into copying, paraphrase classification, and coreference resolution tasks. 

\paragraph{Copying.} 
One of the key skills required for DST that seemed difficult to apply for out-of-domain samples is copying the correct entities mentioned in the conversation to the slot values. 
This skill is relevant to many other natural language understanding tasks that provide multiple candidates that can be chosen for copying, e.g., question  answering and structured text generation such as text-to-SQL. 
To teach better copying skills, we include SQuAD v2.0 \cite{rajpurkar-etal-2018-squadv2}, CoQA \cite{reddy2019coqa}, WikiSQL \cite{zhong2017seq2sql}, and Schema Guided Dialogue (SGD) \cite{rastogi2020towardssgd}.

\paragraph{Paraphrase Classification.}

To internalize an understanding of semantic similarities such that the downstream model become robust to paraphrases, we leverage two paraphrase classification tasks: The Microsoft Research Paraphrase corpus \cite{dolan2005msrp} 
and the Quora Question Pairs corpus \cite{chen2018quora}. 
\paragraph{Coreference Resolution.}

With the expectation that seeing examples that require coreference resolution from other tasks will also help solve cases that need the same skill in DST, we include coreference resolution tasks to our prefinetuning step. We use the Winograd Schema Challenge (WSC) dataset \cite{levesque2012winograd} from the SuperGLUE benchmark \cite{wang2019superglue} and Winograd NLI (WNLI) \cite{ wang-etal-2018-glue}.
/The difference changes the entity that the pronouns in the sentence must resolve to. 

\subsection{Prefinetuning Task Details}
\label{app:prefinetuning_task_details}

The full list of tasks that we use for the prefinetuning step is summarized in \autoref{tab:prefinetuning_tasks}.

\end{document}